\pdfoutput=1

\documentclass[11pt]{article}

\usepackage[]{EMNLP2023}

\usepackage{times}
\usepackage{latexsym}

\usepackage[T1]{fontenc}

\usepackage[utf8]{inputenc}

\usepackage{microtype}

\usepackage{inconsolata}

\usepackage{array}
\newcolumntype{H}{@{}>{\lrbox0}l<{\endlrbox}}

\usepackage{tabularx}
\usepackage{graphics}
\usepackage{graphicx}
\usepackage{float}
\usepackage{multirow}
\usepackage{mathtools}

\usepackage{xcolor,colortbl}
\definecolor{Gray}{gray}{0.85}
\newcolumntype{a}{>{\columncolor{Gray}}c}
\newcolumntype{b}{>{\columncolor{white}}c}
\usepackage{amsmath}
\DeclareMathOperator*{\argmax}{arg\,max}

\newcommand{\method}{\textsc{Cbr-Mrc}}

\newcommand{\answerset}{\ensuremath{\mathcal{A}}}

\newcount\Comments  
\definecolor{darkgreen}{rgb}{0,0.5,0}
\definecolor{darkred}{rgb}{0.7,0,0}
\definecolor{teal}{rgb}{0.1,0.6,0.7}
\definecolor{blue}{rgb}{0.0,0.1,0.9}
\definecolor{orange}{rgb}{1.,0.7,0.0}
\definecolor{lightblue}{rgb}{0.70, 0.80, 0.89}
\definecolor{violet}{rgb}{0.50, 0.16, 0.88}
\newcommand{\kibitz}[2]{\ifnum\Comments=1{{\textcolor{#1}{[#2]}}}\fi}

\definecolor{Gray}{gray}{0.85}
\usepackage{array}
\usepackage{tabularray}
\UseTblrLibrary{booktabs}

\usepackage{tabularx}
\usepackage{graphics}
\usepackage{graphicx}
\usepackage{float}
\usepackage{multirow}
\usepackage{mathtools}
\usepackage{xcolor,colortbl}

\usepackage{amsmath}

\title{Machine Reading Comprehension using Case-based Reasoning}

\author{Dung Thai$^1$, Dhruv Agarwal$^1$, Mudit Chaudhary$^1$, Wenlong Zhao$^1$, Rajarshi Das$^2$\thanks{\ \ Work done while at University of Washington.}\\
\textbf{Manzil Zaheer$^3$, Jay-Yoon Lee$^4$, Hannaneh Hajishirzi$^5$, Andrew McCallum$^1$} \\
        $^1$University of Massachusetts Amherst \\
        $^2$AWS AI Labs, $^3$Google Research, $^4$Seoul National University, $^5$University of Washington \\ 
        \texttt{\{dthai,dagarwal,mchaudhary,wenlongzhao,mccallum\}@cs.umass.edu}}

\begin{document}
\maketitle
\begin{abstract}
We present an accurate and interpretable method for answer extraction in machine reading comprehension that is reminiscent of case-based reasoning (CBR) from classical AI.
Our method (\method{}) builds upon the hypothesis that contextualized answers to similar questions share semantic similarities with each other. 
Given a test question, \method{} first retrieves a set of similar cases from a non-parametric memory and then predicts an answer by selecting the span in the test context that is most similar to the contextualized representations of answers in the retrieved cases.
The semi-parametric nature of our approach allows it to attribute a prediction to the specific set of evidence cases, making it a desirable choice for building reliable and debuggable QA systems.
We show that \method{} provides high accuracy comparable with large reader models and outperforms baselines by 11.5 and 8.4 EM on NaturalQuestions and NewsQA, respectively.
Further, we demonstrate the ability of \method{} in identifying not just the correct answer tokens but also the span with the most relevant supporting evidence.
Lastly, we observe that contexts for certain question types show higher lexical diversity than others and find that \method{} is robust to these variations while performance using fully-parametric methods drops.~\footnote{Our code is available at 
\url{https://github.com/dungtn/cbr-txt}}

\end{abstract}

\section{Introduction}
Machine reading comprehension (MRC) aims to measure the ability of models to understand and reason over a specified sequence of text.
In the extractive setting, the task requires a model to answer a question by reading a set of one or more passages, referred to as the context, and identifying a span of text from that context as the answer.

\begin{figure}[H]
    \centering
\includegraphics[width=\columnwidth]{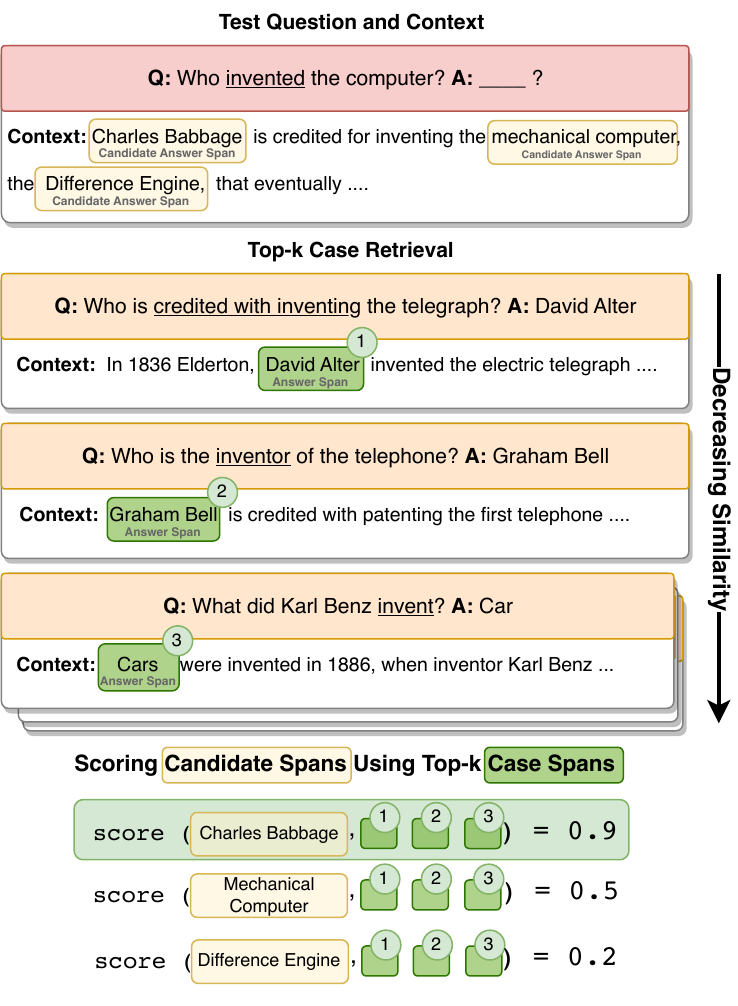} 
    \vspace*{-7mm}
    \caption{\textbf{Overview of \method{}.} \textit{Top:} given a test question, \method{} first retrieves similar cases from a memory of known (question, context, answer) triples. \textit{Bottom:} \method{} scores each candidate answer span by comparing its contextualized representation with the answer spans of the retrieved cases to output a score. The candidate with the highest aggregate score with the answer spans from the cases is output as the prediction (``Charles Babbage'').}
    \label{fig:introfig}
    \vspace*{-3mm}
\end{figure}

Current state-of-the-art machine readers \citep{devlin2019bert, choi2018quac, liu2019roberta, lewis2020rag, izacard2021fid} use fine-tuned transformer-based models \citep{vaswani2017attention}. 
However, the prediction mechanisms in these models are largely opaque to humans, limiting their interpretability and maintainability in real-world scenarios \citep{ribeiro2016why, yang-etal-2018-hotpotqa, camburu2018snli}. 
This becomes particularly relevant when practitioners need to determine the cause of erroneous predictions, for instance, to patch models deployed in production systems. 
Moreover, to build reliable QA systems, it is essential for models to not only predict the correct answer but also provide evidence to support the predictions \cite{thayaparan2020explainability_survey, zeng2020mrc_survey}. 

To address these limitations, we introduce \textbf{\method{}} --- a semi-parametric approach for MRC that makes predictions by explicitly reasoning over a set of retrieved answers for similar questions and their contexts from a non-parametric memory.
Our key idea is that contextualized answers to similar questions share representational similarities. 
We posit that a (question, passage, answer) triple captures a set of \textit{latent relations}, which can be understood to represent the \textit{types} or \textit{intents} present within a question given the accompanying context.
We use the intersection of latent relation sets between questions to learn a similarity function that can then be used to predict answers to questions with similar intents.
More concretely, our approach first compares the contextualized representations of candidate answer spans in the target context with gold answer spans of similar questions from memory. Then, the candidate with the highest aggregate similarity with the gold case answer spans is selected as the predicted answer.
For example, in Figure \ref{fig:introfig}, the question ``\textit{Who invented the computer?}'' shares the same latent relations with the question ``\textit{Who is the inventor of the telephone?}'', and the contextualized span similarity with the case answer ``\textit{Graham Bell}'' is maximized when compared with the correct answer ``\textit{Charles Babbage}'' and low when compared with other candidate spans such as ``\textit{mechanical computer}''.


\method{} adapts the case-based reasoning \citep{schank1983dynamic} paradigm from classical AI \citep{kolodner1983maintaining, rissland1983examples, leake1996case}, which typically consists of four steps --- retrieve, reuse, revise, and retain. 
\method{} uses the first two --- \textbf{retrieval} from a memory of cases, or \textit{casebase}, using the embedding of the test question followed by \textbf{reusing} reasoning patterns encoded in the contextualized answer embeddings of the retrieved cases to select the most similar span from the test context as the predicted answer.
To the best of our knowledge, we are the first to propose a novel adaptation of the CBR framework for the task of question answering over unstructured text.
We empirically verify \method{} on NaturalQuestions and NewsQA, two common MRC datasets, and find our method outperforms strong baselines by up to $11.53$ EM. 
The inference procedure of \method{} is not just accurate but also interpretable --- we know which cases the model uses for its prediction and how much each case contributes.
Further, the semi-parametric nature of \method{} allows for efficient and accurate domain adaptation by simply adding new cases to the casebase without the need to modify any parameters.\footnote{Case augmentation can be extended to make point-fixes for specific inference errors without expensive re-training.}
Lastly, we also find that our method shows higher gains compared to fully-parametric models when the training set contains lexically diverse contexts for questions with similar latent relations, demonstrating the robustness of \method{} when questions may be answered using several different passages.

In summary, our contributions are as follows:
\begin{enumerate}
    \item We propose \method{} --- a novel semi-parametric approach for MRC that outperforms strong baselines on multiple datasets and settings by up to $11.53$ EM on NaturalQuestions and $8.4$ EM on NewsQA.
    \item We show that \method{} makes predictions by identifying the correct supporting evidence. When presented with gold and noisy evidence sentences, \method{} predicts the correct answer span with 83.2 Span-F1 and 74.6 Span-EM compared to BLANC \citep{seonwoo2020blanc} with 79.0 Span-F1 and 66.8 Span-EM on NaturalQuestions.
    \item We show the ability of \method{} for efficient and accurate few-shot domain adaptation. Compared to a strong baseline \citep{friedman2021single}, we achieve a 5.5 EM gain on RelEx \citep{levy-etal-2017-zero} and at-par performance on BioASQ \citep{Tsatsaronis2015AnOO}.
    \item We show that questions with different sets of latent relations can be expressed with varying degrees of lexical diversity in their accompanying contexts and analyze model behavior on such questions. As lexical diversity increases, fully-parametric methods drop by up to 15.4 F1, while \method{} shows robust performance with drops of only up to 4.1 F1. 
    
\end{enumerate}

\begin{figure*}[t]
    \centering
    \hspace{-0.5cm}
    \includegraphics[width=.95\textwidth]{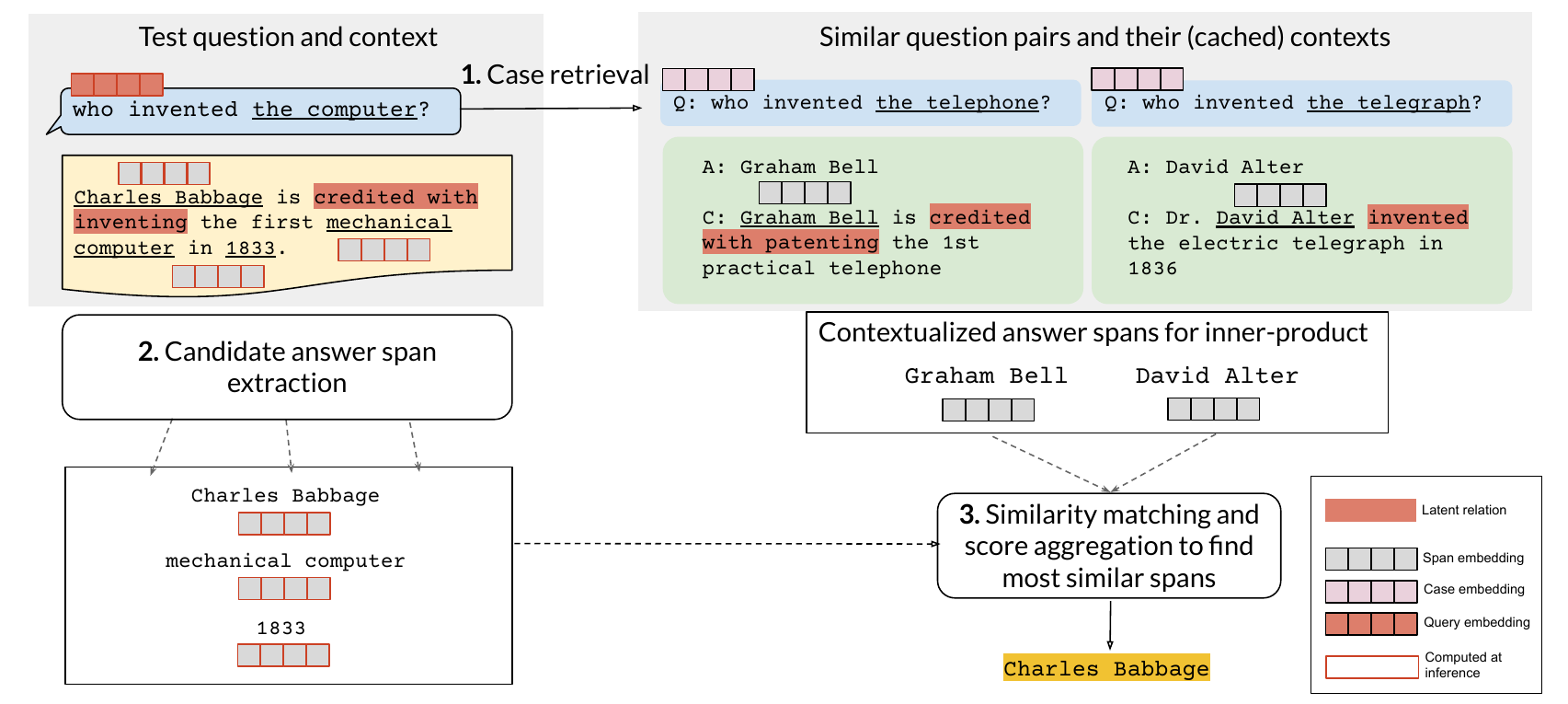}
    \vspace{-3mm}
    \caption{\textbf{Inference with \method.}
    For a given target query, (1) other similar queries (and their contexts) are retrieved; 
    (2) candidate answer spans are extracted from the target context;
    (3) the candidate spans in the target context are ranked w.r.t. answer spans of the retrieved queries by comparing their contextualized embeddings. Finally, the span with the highest inner-product similarity is selected as the prediction. Note that the casebase questions and their corresponding answer span embeddings are pre-computed and cached. 
    }
    \label{fig:concept}
    \vspace{-3mm}
\end{figure*}

\section{Related Work}

\paragraph{Machine Reading Comprehension.} MRC evaluates machine ability to reason over natural language by training it to answer questions on a given passage. Several variations of this task exist, such as cloze-style \cite{https://doi.org/10.48550/arxiv.1506.03340,https://doi.org/10.48550/arxiv.1809.00812}, multiple-choice \cite{richardson-etal-2013-mctest,  lai2017large}, extractive \cite{yang-etal-2015-wikiqa, trischler2016newsqa}, and generative question-answering \cite{https://doi.org/10.48550/arxiv.1611.09268, https://doi.org/10.48550/arxiv.1712.07040}.
Our work focuses on the extractive MRC setting, which aims to identify the correct answer span within a passage given a test question.

Interpretable methods for MRC have also seen much recent interest.  \citet{thayaparan2020explainability_survey} highlights the challenges facing MRC in terms of explainability and its impact on MRC performance. 
Several methods~\citep{https://doi.org/10.48550/arxiv.1412.1632, https://doi.org/10.48550/arxiv.1805.08092, Gravina2018CrossAF} have attempted to improve interpretability in MRC by identifying the supporting sentence for the answer. However, these methods rely on the availability of gold-supporting sentences, which may not always be present. Other methods utilize graph networks~\cite{https://doi.org/10.48550/arxiv.1911.02170,https://doi.org/10.48550/arxiv.1911.00484} and attention mechanisms to capture the attended parts of text or relations \cite{https://doi.org/10.48550/arxiv.1611.01603, Shao_2020}. In this work, we improve interpretability by providing a novel inference procedure that allows us to clearly attribute predictions to previously seen instances that were used in the decision-making process.

\paragraph{Case-based Reasoning.} 
Previous work \cite{cbr_akbc,das-etal-2020-probabilistic,cbr_subg,thai2022cbr, yang2023end} has demonstrated strong performance using CBR on knowledge graph reasoning tasks.
CBR has also been successfully applied to the task of semantic parsing \cite{das2021cbr, pasupat-etal-2021-controllable, awasthi2023structured}, where methods typically retrieve the logical forms of similar cases and pass them through a seq2seq model as input along with a target query.
Both settings, however, operate over structured spaces, 
where similarities between instances can easily be determined by collecting and comparing symbolic patterns in the data, e.g., paths between nodes in a KG, or syntactic similarities in semantic parsing.
In this work, however, we adapt CBR to operate over unstructured text, and it is not obvious how previous approaches can be re-used to work in this setting out-of-the-box.

A concurrent work \cite{iyer-etal-2023-question} performs answer sentence selection by using Graph Neural Networks to model interactions among questions and answers, assuming that for each question-answer, there exist similar question-answers in the training data. 
We instead focus on the more fine-grained task of answer span prediction from contexts and use transformer models to score question-question and answer-answer similarities between each retrieved case and a test example.

\paragraph{In-Context Learning (ICL).} ICL is an emergent property observed in large language models (LLMs) \citep{brown2020language, min2022rethinking} showing remarkable success in several NLP tasks~\citep{cheng2022binding, lewkowycz2022solving, dunn2022structured}. To make predictions, a model is simply prompted with demonstrations for a target task and a test instance. Recent work \citep{rubin2021learning, liu2021makes, li2023few, khattab2022demonstrate} has shown strong performance by retrieving relevant demonstrations. This setting can be understood as a specific instantiation of the case-based reasoning framework, where the ``reuse'' step uses ICL instead of a fine-tuned model. ICL, however, is sensitive to the prompt \citep{chen2022relation}, leading to black-boxed inference and reducing reliability of predictions. Problem decomposition and iterative prompting methods, such as chain-of-thought \citep{wei2022chain} and least-to-most \citep{zhou2022least, drozdov2022compositional}, attempt to address this issue by converting complex problems into intermediate steps. In contrast to methods reliant on emergent pre-training behavior, our work \textit{explicitly} encodes the relational similarity hypothesis in a trained model to reuse exemplars for making predictions.

\section{Method}
\label{sec:method}

We now formally describe \method{} (Figure \ref{fig:concept}).
A \textit{case} in CBR consists of a problem and its solution. 
Formally, we define a case $c$ as a tuple $\{q, \answerset, p\}$, representing a question, its corresponding answer set\footnote{$\answerset$ is a singleton if every question has a unique answer.}, and the passage context that contains each element in the answer set. The context $p$ is a span of text that not only contains all $a \in \answerset$ but also captures one or more latent relations between the question and the answer. 
We refer to a collection of all cases as the \textit{casebase} $\mathcal{B}$, which is typically set to the full training data and may continually be augmented to incorporate new data patterns. 
In the MRC setting, annotations for $p$ for all $c \in \mathcal{B}$ are assumed to be available. 

\subsection{Case Retrieval}
Given an input question $q$, \method{} retrieves \emph{similar} cases from the casebase, where similarity is determined by the overlap of latent relations encoded by the query and the casebase questions.
Formally, let $\mathcal{R}_q$ be the set of latent relations expressed in $q$.
For example, for the question ``\textit{who invented the telephone?}'', the latent relation expressed is the concept ``\textit{invented by}''.\footnote{Note that \method{} does not require explicit annotations of question relation types, e.g., mapping to KB schema relations. It instead leverages such semantics implicitly.}
A case $c_k \coloneqq \{q_k, \answerset_k, p_k\}$ is similar to $q$, if the latent relations expressed in $q$ and $q_k$ overlap, i.e., $\mathcal{R}_q \cap \mathcal{R}_{q_k} \neq \emptyset$. For the above example, a similar question may be ``\textit{Who was the inventor of the television?}''. 
We hypothesize that these latent relations are captured in dense representations obtained from language models; thus, relational overlap can be assessed via vector similarity scores.
In our experiments, we use a pre-trained BERT \citep{devlin2019bert} model to encode questions and use the representation of the \texttt{[CLS]} token for similarity calculations.

To ensure that a question representation faithfully captures the expressed relations and is not influenced by the entities in the question, we replace mentions of all entities with a \texttt{[MASK]} token, \textit{e.g.,} ``\textit{who invented the telephone?}'' becomes ``\textit{who invented the} \texttt{[MASK]}\textit{?}''. 
Previous work~\cite{soares2019matching} has shown this technique to be useful in learning entity-independent relational representations.
In \method{}, masking removes spurious similarities, where the presence of a common entity results in two relationally divergent questions being adjudged as similar, for e.g., ``\textit{who invented the telephone?}'' and ``\textit{what is the telephone used for?}'' should be considered dissimilar in our setup. In practice, we use T-NER \cite{tner}, an automatic entity recognizer, to identify mentions of all entities in the question.

The similarity score between two masked queries is computed as the inner product between their normalized vector representations (cosine similarity). 
The representations for all questions in the casebase are pre-computed and cached. During inference, we retrieve the top-$k$ similar cases for a query by performing a fast nearest-neighbor search over the cached representations.

\paragraph{Case retrieval during training.} 
To minimize the number of retrieved cases with a low overlap in latent relations during training, we additionally employ two filtering heuristics --- 
(1) we use a minimum threshold of 0.95 for the cosine similarity score to filter out dissimilar cases, and
(2) we use a \emph{wh-filter} to remove retrieved cases where the question keyword (e.g., \textit{"who"}, \textit{"where"}, \textit{"when"}) present in the case question does not match the keyword present in the input query.

\subsection{Case Reuse}

We hypothesize that contexts associated with similar questions also share relational similarities. 
For instance, Figure~\ref{fig:concept} shows the context surrounding the answer span of the retrieved case  ``\textit{\textbf{Graham Bell} is \underline{credited with patenting}...}'' and the target context surrounding the gold answer ``\textit{\textbf{Charles Babbage} is \underline{credited with inventing}...}''
In \method{}, we use the contextualized embeddings of the answer spans from retrieved cases to find this semantic overlap within the target context. 
For each candidate answer span in the context, a similarity score is explicitly computed with answer representations from the retrieved cases. 

For efficient processing, we employ a dense retriever that uses contextualized answer spans from past cases as queries and searches for the target answer over the input passage embeddings.
Given an input question $q$ and an input passage $p$, we search for a text span $s \in \mathcal{S}_p$, the set of all possible spans in $p$, whose representation is the most similar to representations of answer spans in similar question contexts.
We employ a standard BERT-base model as the passage encoder of the dense retriever.
Due to the combinatorial complexity of considering all $s \in \mathcal{S}_p$, we follow \citet{denspi} and consider only a subset of candidate spans in $p$ likely to be the answer. 
In particular, we consider a set of candidates $\mathcal{S}_\text{cand} \subseteq \mathcal{S}_p$ such that $\mathcal{S}_\text{cand}$ is composed of only entity mentions, date-time strings, numbers, and quoted strings that can be extracted from $p$.
To increase span coverage, we include all n-grams up to length three.
See \ref{app:reuse-candidate-spans} for more details regarding the candidate span extraction process. 

We now formulate the task to predict the answer $\hat{a}$ to the question $q$ from a context passage $p$ given its relevant past cases.
Let $c_k \coloneqq \lbrace q_k, \answerset_k, p_{k}\rbrace$ be the $k$-th retrieved case for the given question $q$. 
For an answer $a_k \in \answerset_k$, let $\mathtt{Enc}(a_k)$ be the contextualized answer span embedding pooled from the embeddings of the answer tokens in $p_{k}$.
Similarly, for all candidate answers $s \in \mathcal{S}_\text{cand}$, let $\mathtt{Enc}(s)$ be their contextualized embeddings pooled from the encoded tokens in context $p$.
The predicted answer span based on the $k$-th retrieved case is given by 
\begin{align}
    \hat{a}^{(k)} = \argmax_{s \in \mathcal{S}_\text{cand}}\max_{a_k \in \answerset_k} \mathtt{Enc}(s)^T\mathtt{Enc}(a_k), \nonumber
\end{align}
i.e., the candidate answer span most similar to the answer set $\answerset_k$ based on similarity scores computed using inner products.
At inference, the similarity scores of each candidate span are aggregated across the top-$k$ retrieved cases by taking the maximum scores. 
The candidate span with the highest aggregated score is predicted as the answer. 


\subsubsection{Training}
The passage encoder must be fine-tuned to maximize similarity scores between answer span embeddings of similar questions. We fine-tune a pre-trained BERT model with a contrastive loss \citep{khosla2020supervised} function using answer and non-answer spans.

Specifically, consider a train question $q_t$, its set of answers $\answerset_t$, and its set of candidate spans $\mathcal{S}_{\text{cand},t} \subseteq \mathcal{S}_{p_t}$ derived from the passage $p_t$. We use the gold answer set $\answerset_t$ as the set of positive candidate spans, and define the set of negative candidate spans as $\mathcal{S}_{\text{cand},t} \setminus \answerset_t$.
Let $\mathrm{sim}(\cdot, \cdot) \coloneqq \mathtt{Enc}(\cdot)^T\mathtt{Enc}(\cdot)$ be the similarity function (dot product or cosine similarity) between vector representations.
For the top-$k$ similar questions as described in the previous section, we fine-tune the passage encoder with the following objective derived from the soft nearest-neighbor loss \citep{salakhutdinov2007snn,frosst2019snn}:
\begin{align*}
    \mathcal{L}_t &= - \log\sum_{s \in \answerset_t} \max_k \max_{a_k \in \answerset_k} \exp(\mathrm{sim}(a_k, s)/\tau) \nonumber \\
    & \hspace{-0.5cm}+ \log\sum_{s \in \mathcal{S}_{\text{cand},t} \setminus \answerset_t} \max_k \max_{a_k \in \answerset_k} \exp(\mathrm{sim}(a_k, s)/\tau), \label{eq:loss}
\end{align*}%
where $\tau$ is a temperature hyperparameter. The loss for the entire training set is simply the mean of $\mathcal{L}_t$ over all $q_t \in \mathcal{Q}$, the training set.
Among the top-$k$ retrieved cases, we pick the spans with the highest similarity scores.
This loss function encourages the similarity scores of the gold answer spans with respect to the answers in the top-$k$ cases to be higher than the similarity of other spans of texts in the input context that are not the answer.

\section{Experiments}



We evaluate \method{} on two machine reading comprehension tasks --- supervised answer span extraction and few-shot domain adaptation. 
For three MRC datasets, we use the training data to both construct the casebase as well as train our model and report performance on the in-domain test sets. We report the few-shot domain adaptation performance on two additional datasets.

\subsection{Datasets}

\textbf{NaturalQuestions (NaturalQ)} consists of over 323K questions and answers, representative of real queries people make on the Google search engine \cite{47761}. 
The answers are sourced from Wikipedia and annotated by crowd workers, including both a long and short version. Our experiments utilize short answers as the gold answers and use the long answers as context.
\textbf{NewsQA} has 120K crowd-sourced questions based on CNN news articles \cite{trischler2016newsqa}. 
Annotation is obtained by two sets of crowd workers: one creating questions based on article highlights and the other answering them using spans from the full article. 
\textbf{BioASQ} is a large-scale biomedical dataset that features question-answer pairs crafted by domain experts \cite{Tsatsaronis2015AnOO}. 
Questions are linked to relevant scientific literature found on PubMed, with the abstracts of these articles serving as the context for MRC.
\textbf{RelationExtraction (RelEx)} comprises question-answer pairs that express Wikidata relations between entities mentioned in Wikipedia articles \cite{levy-etal-2017-zero}. 
The dataset is derived from the WikiReading \citep{hewlett2016wikireading} slot-filling dataset through predefined templates.

We utilize the curated version of all the previously mentioned datasets from the MRQA shared task \citep{fisch2019mrqa}, including annotations for the gold answer spans within their context. 
The gold context for constructing a case, as described in Section \ref{sec:method}, is a short paragraph surrounding the gold answer span taken from the originally provided text. Additionally, note that we use BioASQ and RelEx to \emph{only evaluate} in the few-shot setting and not train our models.




\subsection{Experiment Setup}
We refer the reader to Appendix~\ref{app:exp-implementation} for more details on the models used in this work.

\paragraph{Baselines.} 
We compare \method{} to BERT~\citep{devlin2019bert}, RoBERTa~\citep{liu2019roberta}, ALBERT~\citep{DBLP:journals/corr/abs-1909-11942}, and SpanBERT~\citep{joshi-etal-2020-spanbert} as well as the SOTA baselines for our settings -- BLANC~\citep{seonwoo-etal-2020-context} and MADE~\citep{friedman2021single}. 

\paragraph{Evaluation Metrics.\footnote{(1) All reported numbers are from single runs on the test set unless noted otherwise. (2) Hyperparameters used by our models were tuned on a development set and the details are included in the appendix.}}
We report exact match (EM) and F1 scores to measure model performance. 
EM calculates the percentage of predictions that match all tokens of any one of the correct answers for a given question. 
F1 score measures the token overlap between the prediction and the correct answer.
As these metrics do not take into account the surrounding context of the answer span (i.e., the supporting evidence), we also report Span-EM and Span-F1, following \citet{seonwoo-etal-2020-context}, which consider overlapping indices between the predicted and gold answer spans and exclude answers with irrelevant supporting text.



\begin{table}[t]
\centering
\resizebox{0.9\columnwidth}{!}{
    \begin{tabular}{l c c c c}
    \toprule
    & \multicolumn{2}{c}{\textbf{NewsQA}} & \multicolumn{2}{c}{\textbf{NaturalQ}} \\
    \cmidrule(lr){2-5}
    \bf Model & \bf EM & \bf F1 & \bf EM & \bf F1 \\
    \midrule 
    BERT & 50.11 & 65.07 & 64.48 & 76.39 \\
    ALBERT & 51.18 & 66.02 & 63.81 & 75.89 \\
    RoBERTa & 52.36 & 67.28 & 66.33 & 78.54 \\
    MADE & \underline{56.55} & \textbf{72.12} & 68.86 & 80.09 \\
    SpanBERT & 52.85 & 67.93 & 66.60 & 78.31 \\
    BLANC & 55.52 & \underline{70.31} & 68.33 & 80.04 \\
    SpanBERT$_{\textsc{large}}$ & 53.84 & 69.06 & 69.14 & 80.66 \\
    BLANC$_{\textsc{large}}$ & 57.40 & 72.36 & \underline{70.59} & \underline{81.99} \\
    \midrule
    \textbf{\method{}} & \textbf{64.95} & 69.17 & \textbf{82.12} & \textbf{83.10} \\
    \bottomrule
    \end{tabular}
}
\vspace{-1mm}
\caption{\textbf{MRC performance on NewsQA and NaturalQ.}  \method{} outperforms all baselines in finding the exact match of the answers. On NewsQA, our model shows +8.4 EM over MADE but has lower F1 which indicates \method{} focuses on getting the right answer rather than overfitting to the answer tokens.}
\label{tab:main_results}
\vspace{-5mm}
\end{table}



\subsection{Results}
\label{sec:main_results}
We report our main MRC results on NewsQA and NaturalQ in Table \ref{tab:main_results}.
We find that \method{} consistently outperforms all baselines. 
On NaturalQ, \method{} achieves the state-of-the-art, outperforming the next best model (BLANC$_{\textsc{large}}$) by 11.53 EM. 
Similarly, on NewsQA, \method{} shows an improvement of 12.56 EM over BLANC$_{\textsc{large}}$, and 8.4 EM over the next best model (MADE).
Note that the best baseline model (BLANC$_{\textsc{large}}$) has three times the number of parameters as \method{}. 
Furthermore, our model shows performance gains in EM but a lower F1 which indicates \method{} focuses on getting the correct answer rather than over-fitting to the answer tokens.

\subsubsection{Supporting Evidence Identification}
It is becoming increasingly important to train models to predict the \textit{right} answer for the \textit{right} reasons. Since \method{} identifies the answer span by comparing it with multiple answer-containing contexts of similar cases, we hypothesize that it can cut through the noise and identify the correct occurrence of the span that \textit{exactly} answers the question. Our main baseline is BLANC~\cite{seonwoo-etal-2020-context}, which trains an auxiliary model specifically for context prediction.

Table \ref{tab:span_results} highlights the performance gains of \method{} compared to BLANC and several strong baselines. Our method shows improvements on NaturalQ by 7.89 Span-EM and 4.12 Span-F1 points in predicting the span indices. On NewsQA, it improves performance by 1.68 Span-EM and 1.73 Span-F1 points compared to BLANC. To determine the gains due to predicting the answer span with the correct supporting evidence, we follow the approach in BLANC and evaluate performance on the NaturalQuestions (NaturalQ*) subset that contains questions with answers that have at least two mentions within the context. Our method outperforms BLANC on this subset by 6.73 and 4.05 points for Span-EM and Span-F1, respectively.

\begin{table}[t]
    \centering
    \resizebox{\columnwidth}{!}{
        \begin{tabular}{llcc}
            \toprule
            \textbf{Dataset} & \textbf{Model} & \textbf{Span-EM}& \textbf{Span-F1} \\
            \midrule
                \multirow{7}{*}{NaturalQ}         & BERT                            & 60.63                           & 72.92                           \\
                                      & ALBERT                          & 60.31                           & 72.66                           \\
                                      & RoBERTa                         & 62.59                           & 75.07                           \\
                                      & SpanBERT                        & 62.71                           & 75.16                           \\
                                      & BLANC                           & 64.57                           & 76.99                           \\ 
                                      & SpanBERT$_{\textsc{large}}$               & 65.28                           & 77.62                           \\
                                      & BLANC$_{\textsc{large}}$                  & 66.75                           & 79.04                           \\ \cmidrule(lr){2-4} 
                                      & \textbf{\method{}}      & \textbf{74.64}                  & \textbf{83.16}                  \\ \midrule
    \multirow{7}{*}{NewsQA}           & BERT                            & 45.53                           & 59.18                           \\
                                      & ALBERT                          & 46.54                           & 60.12                           \\
                                      & RoBERTa                         & 47.43                           & 61.36                           \\
                                      & SpanBERT                        & 48.04                           & 62.26                           \\
                                      & BLANC                           & 50.60                           & 64.39                           \\ 
                                      & SpanBERT$_{\textsc{large}}$             & 49.03                           & 63.43                           \\
                                      & BLANC$_{\textsc{large}}$                  & 52.39                           & 66.48                           \\ \cmidrule(lr){2-4}
                                      & \textbf{\method{}}      & \textbf{54.07}                  & \textbf{68.21}                  \\ \midrule
    \multirow{4}{*}{NaturalQ*}        & RoBERTa                         & 60.12                           & 65.99                           \\
                                      & SpanBERT                        & 57.63                           & 63.47                           \\
                                      & BLANC                           & 61.43                           & 67.07                           \\ \cmidrule(lr){2-4} 
                                      & \textbf{\method{}}      & \textbf{71.30}                  & \textbf{81.04}                  \\ 
            \bottomrule
        \end{tabular}
     }
    \caption{\textbf{Identifying the correct supporting evidence.} We report Span-F1 and Span-EM scores for \method{} and baseline models, indicating how often the span with the correct supporting evidence is identified versus selecting either the incorrect answer or a spurious occurrence of the correct  answer. We report baseline numbers for BLANC as reported in their paper. NaturalQ* is a subset of NaturalQuestions containing questions with answers having at least two mentions within the context.}
    \label{tab:span_results}
    \vspace{-3mm}    
\end{table}

\begin{figure}[t]
    \centering
    \hspace*{-0.5cm}
    \includegraphics[width=\columnwidth]{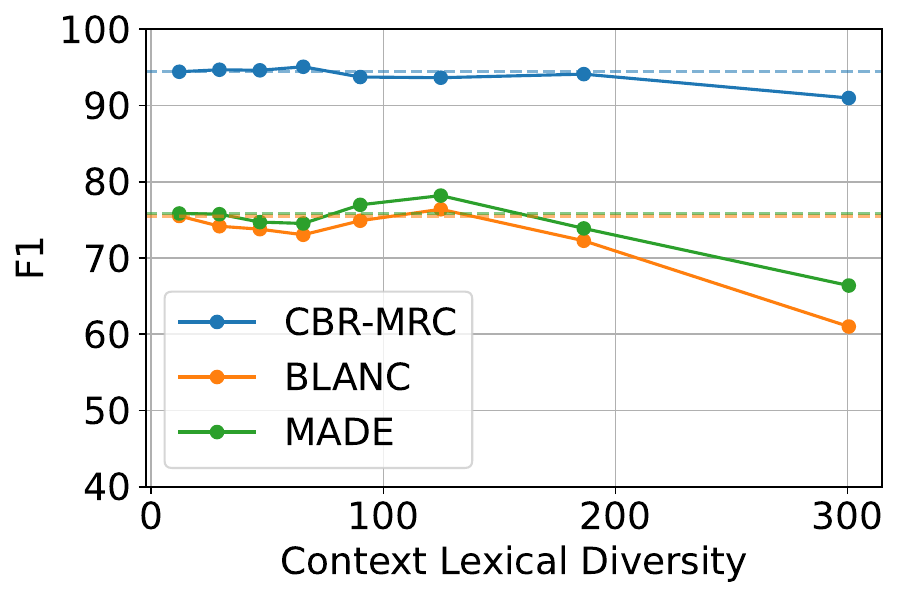}
    \vspace{-2mm}
    \caption{\textbf{Robustness to lexical diversity in passages.} Lexical diversity of latent relation clusters is the number of unique tokens in passages seen at training for that latent relation cluster (or question type). We plot F1 performance (averaged over 6 clusterings varying cut thresholds) on 8 buckets of increasing lexical diversity for the latent relations seen at training. \method{} shows a drop in performance only up to 4.10 points, while BLANC and MADE show drops of up to 15.38 and 11.80 points, respectively.}
    \label{fig:lexical-diversity}
    \vspace{-3mm}
\end{figure}
\subsubsection{Robustness to Lexical Diversity}

\method\ performs answer extraction 
by explicitly comparing span similarities of the target context with the context spans in the filtered casebase. 
Purely parametric methods can also be seen as following a similar mechanism, albeit performing the two steps implicitly. 
These models, therefore, must rely only on the parameter values learned at training in order to handle several relational semantics, which may each be expressed in contexts with varying levels of diversity in language form. In practice, we find that some relations are indeed expressed with more lexical diversity than others (Appendix \ref{app:lex-NQ}). In Figure \ref{fig:lexical-diversity}, we investigate how \method{} performs at inference under varying amounts of lexical diversity observed in passage contexts at \textit{training}. We first cluster questions according to their embeddings and then compute n-gram statistics \citep{li2015diversity, duvsek2020evaluating, tevet2020evaluating} across passage contexts found in each cluster to measure diversity. We use NaturalQuestions as a case study for our analysis and compare performance with BLANC and MADE, two fully-parametric baselines.
\paragraph{Clustering questions by latent relations.}  We cluster questions in the training set based on the latent relations they encode using hierarchical agglomerative clustering (HAC) \citep{krishna2012efficient} over similarity scores. For a fair evaluation, we compute 6 different flat clusterings of varying cluster tightness, i.e., how similar each question is within a cluster. For each clustering, we compute the per-cluster lexical diversity as the count of the number of unique tokens in the set of passage contexts across questions in the cluster, normalized by the cluster size. We then cluster the test set by assigning each question to the cluster label of its top-1 similar train question for each of the 6 clusterings, resulting in 6 \textit{test} clusterings (see Appendix~\ref{app:lex-clustering} for more details).

\paragraph{Results.} Figure \ref{fig:lexical-diversity} shows the mean F1 scores between \method{} and the baselines on test queries for varying values of lexical diversity observed at training for the same latent relations as the test queries. 
We group clusters by lexical diversity scores into buckets and report the aggregated F1 scores for each bucket.\footnote{(1) We use $k$-means over lexical diversity scores of all clusters to construct $B=8$ buckets. (2) We report the F1 performance on all 6 clusterings in Appendix \ref{app:lex-plots}.}
On average, the performance gap between \method{} and baselines tends to increase as lexical diversity increases. The min-max F1 difference, which captures the drop in performance as lexical diversity increases, is $4.10$ points for \method{} while BLANC and MADE show significantly larger differences of $15.38$ and $11.80$ points, respectively.
We posit that as diversity in contexts for the same latent relation increases, it is harder for fully parametric models to learn parameters that are suitable for all lexical variations. On the other hand, the semi-parametric nature of \method\ allows it to partially shift this burden to test time and ground itself using only a few retrieved cases that are explicitly compared.


\begin{table}[t]
\centering
\resizebox{0.8\columnwidth}{!}{
    \begin{tabular}{lcccc}
    \toprule 
    & \multicolumn{2}{c}{\textbf{BioASQ}} & \multicolumn{2}{c}{\textbf{RelEx}} \\
    \cmidrule(lr){2-5}
    \bf \textbf{Model}  & \bf EM & \bf F1 & \bf EM & \bf F1 \\
    \midrule 
    MADE$_{\textsc{zero}}$ & $43.3$ & $62.9$ & $72.0$ & $84.9$ \\
    MADE$_{\textsc{ft}}$ & $\mathbf{62.5}$ & $\mathbf{74.5}$ & $82.3$ & $\mathbf{90.0}$ \\
    \midrule
    \textbf{\method{}} & $62.4$ & $64.6$ & $\mathbf{87.8}$ & $88.2$ \\
    \bottomrule
    \end{tabular}
}
\vspace{-1mm}
\caption{\textbf{Few-shot domain adaptation.} We compare \method{} with two variants of the MADE adapter networks, which are also trained on NaturalQuestions. MADE$_{\textsc{zero}}$ demonstrates zero-shot performance, while MADE$_{\textsc{ft}}$ is further fine-tuned on $256$ samples for each dataset. \method{} is evaluated by simply adding the samples into the casebase without any fine-tuning.}
\label{tab:zeroshot_results}
\vspace{-3mm}
\end{table}


\subsubsection{Few-shot Domain Adaptation}
It is important for models to generalize to unseen datasets. 
We test \method{} trained on NaturalQ on two diverse out-of-domain datasets - BioASQ for MRC ~\cite{Tsatsaronis2015AnOO} and RelEx for relation extraction ~\citep{levy-etal-2017-zero}.
We compare our model with the MADE adapter networks \citep{friedman2021single} that are trained on NaturalQuestions, and further fine-tuned on $256$ samples from the target dataset.
We follow the same practice in \citep{friedman2021single} and evaluate \method{} on a held-out subset of $400$ samples. 
For both datasets, we use the trained model obtained from the experiment on the NaturalQuestions dataset, and the casebase for each dataset is built from the same training set that MADE also uses. 

Table \ref{tab:zeroshot_results} presents the performance of \method{} and MADE adapters.
Our model outperforms the zero-shot adapter network (MADE$_{\textsc{zero}}$) by simply putting some samples from the target dataset into the casebase. 
When compared to the fine-tuned adapter (MADE$_{\textsc{ft}}$), our model gains $5.5$ EM points on RelEx while being comparable to MADE$_{\textsc{ft}}$ on BioASQ. 
These results confirm that \method{} can transfer to a new domain by simply collecting a handful of sample cases, which is significantly less expensive than model fine-tuning. 

We notice a drop in F1 scores on both datasets. F1 scores evaluate the partial matching of answers. The inference procedure used by previous work, including our baselines, independently predicts the start and end spans of the answer, which encourages the inclusion of partial matches. In \method{}, on the other hand, the entire candidate span is considered when making a prediction by comparing against the representations of gold spans from the casebase. This results in higher similarity scores for candidate spans that map to the same answer type, rather than matching the correct answer partially, which explains lower F1 scores for \method{}. 

\begin{table}[t]
\centering
\resizebox{0.8\columnwidth}{!}{
    \begin{tabular}{cHcccc}
    \toprule 
     \bf Cases ($k$) & $\bf k=0$ & $\bf 1$ & $\bf 5$ & $\bf 10$ & $\bf 20$ \\
    \midrule
    \bf EM & $00$ & $82.03$ & $\bf 82.12$ & $77.95$ & $77.39$ \\
    \bf F1 & $00$ & $82.13$ & $\bf 83.10$ & $78.00$ & $78.45$ \\
    \bottomrule
    \end{tabular}
}
\vspace{-1mm}
\caption{\textbf{Performance with different number of case retrievals $k \in \{1, 5, 10, 20\}$.} Both EM and F1 display the same trend. Both performance metrics increase as $k$ increases from 1 to 5. However, larger sets of retrievals ($k > 10$) reduce performance to levels worse than even $k=1$, which indicates the introduction of noise. 
}
\label{tab:k_results}
\vspace{-3mm}
\end{table}


\subsubsection{Effect of Retrieval Quantity}

We explore how the number of retrieved cases $k$ affects the reading comprehension performance, via varying $k$ on the NaturalQuestions dataset.
We find that while increasing $k$ does enrich the set of contexts for referencing in the reuse step, too many retrievals may introduce irrelevant cases that degrade performance.

\subsubsection{Effect of Modeling}
We analyze CBR-MRC performance with different base models and show that CBR-MRC can improve upon them. 
We run the experiments on NQ with the same set of hyper-parameters. 

\begin{table}[h]
\centering
\resizebox{\columnwidth}{!}{
    \begin{tabular}{lcccc}
    \toprule 
    \bf \textbf{Model}  & \bf EM & \bf F1 & \bf Span-EM & \bf Span-F1 \\
    \midrule 
    BERT & $64.48$ & $76.39$ & $60.63$ & $72.92$ \\
    \textbf{\method{}}$_{\textsc{BERT}}$ & $\mathbf{82.12}$ & $\mathbf{83.1}$ & $74.64$ & $\mathbf{83.16}$ \\
    \midrule
    DeBERTa & $43.11$ & $74.07$ & $36.03$  & $63.7$ \\
    \textbf{\method{}}$_{\textsc{DeBERTa}}$ & $80.30$ & $82.22$ & $\mathbf{76.09}$ & $80.39$ \\
    \bottomrule
    \end{tabular}
}
\vspace{-2mm}
\caption{\textbf{Performances of \method{} variants on NQ.} We compare two variants of \method{} with BERT and DeBERTa as its base model.}
\label{tab:ablation-model}
\vspace{-3mm}
\end{table}









\section{Conclusion}
We present \method{}, a semi-parametric model for machine reading comprehension that is simple, accurate, and interpretable. Our model stores a collection of cases, retrieves the most relevant cases for a given test question, and then explicitly reuses the reasoning patterns encoded in the embeddings of these cases to predict an answer.
We show that our model performs well for both extracting answers and identifying supporting evidence on several MRC tasks compared to fully-parametric baselines. We also demonstrate the ability of our model to transfer to new domains with limited labeled data.
Finally, we analyze our model under varying conditions of lexical diversity and find that it is robust to high lexical diversity, whereas fully-parametric models show a drop in performance.



\section{Limitations}
\label{sec:limitations}

\method{}, and the CBR framework more generally, relies on the existence of past cases to make predictions for a new problem. 
This can pose a challenge for composite questions, which require multi-hop reasoning, since the likelihood of enumerating each algebraic combination of reasoning patterns in the casebase is impractical.
Models, thus, may only be able to match a portion of the question, resulting in partially correct reasoning. 
To address these limitations, future work could explore methods to explicitly encourage such compositional generalization, such as question decomposition with a recursive application of CBR.

\section*{Ethics Statement}

The objective of our research and the proposed methodology is to enhance performance on the Machine Reading Comprehension task. The datasets utilized in this study have been extensively employed in previous research and, as far as our knowledge extends, are not associated with any privacy or ethical concerns.

\section*{Acknowledgements}
\label{acknowledgements}
We thank members of UMass IESL for helpful discussions and feedback. 
This work was supported in part by Amazon Digital Services, in part by IBM Research AI through the AI Horizons Network, in part by the Chan Zuckerberg Initiative under the project "Scientific Knowledge Base Construction", and in part using high-performance computing equipment obtained under a grant from the Collaborative R\&D Fund managed by the Massachusetts Technology Collaborative. Any opinions, findings, and conclusions / recommendations expressed in this material are those of the authors and do not necessarily reflect those of the sponsor(s).

\bibliography{anthology,custom}

\begin{thebibliography}{73}
\expandafter\ifx\csname natexlab\endcsname\relax\def\natexlab#1{#1}\fi

\bibitem[{Awasthi et~al.(2023)Awasthi, Chakrabarti, and
  Sarawagi}]{awasthi2023structured}
Abhijeet Awasthi, Soumen Chakrabarti, and Sunita Sarawagi. 2023.
\newblock Structured case-based reasoning for inference-time adaptation of
  text-to-sql parsers.
\newblock \emph{arXiv preprint arXiv:2301.04110}.

\bibitem[{Bajaj et~al.(2016)Bajaj, Campos, Craswell, Deng, Gao, Liu, Majumder,
  McNamara, Mitra, Nguyen, Rosenberg, Song, Stoica, Tiwary, and
  Wang}]{https://doi.org/10.48550/arxiv.1611.09268}
Payal Bajaj, Daniel Campos, Nick Craswell, Li~Deng, Jianfeng Gao, Xiaodong Liu,
  Rangan Majumder, Andrew McNamara, Bhaskar Mitra, Tri Nguyen, Mir Rosenberg,
  Xia Song, Alina Stoica, Saurabh Tiwary, and Tong Wang. 2016.
\newblock \href {https://doi.org/10.48550/ARXIV.1611.09268} {Ms marco: A human
  generated machine reading comprehension dataset}.

\bibitem[{Brown et~al.(2020)Brown, Mann, Ryder, Subbiah, Kaplan, Dhariwal,
  Neelakantan, Shyam, Sastry, Askell et~al.}]{brown2020language}
Tom Brown, Benjamin Mann, Nick Ryder, Melanie Subbiah, Jared~D Kaplan, Prafulla
  Dhariwal, Arvind Neelakantan, Pranav Shyam, Girish Sastry, Amanda Askell,
  et~al. 2020.
\newblock Language models are few-shot learners.
\newblock \emph{Advances in neural information processing systems},
  33:1877--1901.

\bibitem[{Camburu et~al.(2018)Camburu, Rockt{\"a}schel, Lukasiewicz, and
  Blunsom}]{camburu2018snli}
Oana-Maria Camburu, Tim Rockt{\"a}schel, Thomas Lukasiewicz, and Phil Blunsom.
  2018.
\newblock e-snli: Natural language inference with natural language
  explanations.
\newblock \emph{Advances in Neural Information Processing Systems}, 31.

\bibitem[{Chen et~al.(2022)Chen, Zhao, Yu, McKeown, and He}]{chen2022relation}
Yanda Chen, Chen Zhao, Zhou Yu, Kathleen McKeown, and He~He. 2022.
\newblock On the relation between sensitivity and accuracy in in-context
  learning.
\newblock \emph{arXiv preprint arXiv:2209.07661}.

\bibitem[{Cheng et~al.(2022)Cheng, Xie, Shi, Li, Nadkarni, Hu, Xiong, Radev,
  Ostendorf, Zettlemoyer et~al.}]{cheng2022binding}
Zhoujun Cheng, Tianbao Xie, Peng Shi, Chengzu Li, Rahul Nadkarni, Yushi Hu,
  Caiming Xiong, Dragomir Radev, Mari Ostendorf, Luke Zettlemoyer, et~al. 2022.
\newblock Binding language models in symbolic languages.
\newblock \emph{arXiv preprint arXiv:2210.02875}.

\bibitem[{Choi et~al.(2018)Choi, He, Iyyer, Yatskar, Yih, Choi, Liang, and
  Zettlemoyer}]{choi2018quac}
Eunsol Choi, He~He, Mohit Iyyer, Mark Yatskar, Wen-tau Yih, Yejin Choi, Percy
  Liang, and Luke Zettlemoyer. 2018.
\newblock Quac: Question answering in context.
\newblock \emph{arXiv preprint arXiv:1808.07036}.

\bibitem[{Das et~al.(2020{\natexlab{a}})Das, Godbole, Dhuliawala, Zaheer, and
  McCallum}]{cbr_akbc}
Rajarshi Das, Ameya Godbole, Shehzaad Dhuliawala, Manzil Zaheer, and Andrew
  McCallum. 2020{\natexlab{a}}.
\newblock A simple approach to case-based reasoning in knowledge bases.
\newblock In \emph{AKBC}.

\bibitem[{Das et~al.(2020{\natexlab{b}})Das, Godbole, Monath, Zaheer, and
  McCallum}]{das-etal-2020-probabilistic}
Rajarshi Das, Ameya Godbole, Nicholas Monath, Manzil Zaheer, and Andrew
  McCallum. 2020{\natexlab{b}}.
\newblock \href {https://doi.org/10.18653/v1/2020.findings-emnlp.427}
  {Probabilistic case-based reasoning for open-world knowledge graph
  completion}.
\newblock In \emph{Findings of the Association for Computational Linguistics:
  EMNLP 2020}, pages 4752--4765, Online. Association for Computational
  Linguistics.

\bibitem[{Das et~al.(2022)Das, Godbole, Naik, Tower, Jia, Zaheer, Hajishirzi,
  and McCallum}]{cbr_subg}
Rajarshi Das, Ameya Godbole, Ankita Naik, Elliot Tower, Robin Jia, Manzil
  Zaheer, Hannaneh Hajishirzi, and Andrew McCallum. 2022.
\newblock Knowledge base question answering by case-based reasoning over
  subgraphs.
\newblock In \emph{ICML}.

\bibitem[{Das et~al.(2021)Das, Zaheer, Thai, Godbole, Perez, Lee, Tan,
  Polymenakos, and McCallum}]{das2021cbr}
Rajarshi Das, Manzil Zaheer, Dung~Ngoc Thai, Ameya Godbole, Ethan Perez,
  Jay~Yoon Lee, Lizhen Tan, Lazaros Polymenakos, and Andrew McCallum. 2021.
\newblock Case-based reasoning for natural language queries over knowledge
  bases.
\newblock In \emph{EMNLP}.

\bibitem[{Devlin et~al.(2019)Devlin, Chang, Lee, and
  Toutanova}]{devlin2019bert}
Jacob Devlin, Ming-Wei Chang, Kenton Lee, and Kristina Toutanova. 2019.
\newblock Bert: Pre-training of deep bidirectional transformers for language
  understanding.
\newblock In \emph{NAACL}.

\bibitem[{Drozdov et~al.(2022)Drozdov, Sch{\"a}rli, Aky{\"u}rek, Scales, Song,
  Chen, Bousquet, and Zhou}]{drozdov2022compositional}
Andrew Drozdov, Nathanael Sch{\"a}rli, Ekin Aky{\"u}rek, Nathan Scales, Xinying
  Song, Xinyun Chen, Olivier Bousquet, and Denny Zhou. 2022.
\newblock Compositional semantic parsing with large language models.
\newblock \emph{arXiv preprint arXiv:2209.15003}.

\bibitem[{Dunn et~al.(2022)Dunn, Dagdelen, Walker, Lee, Rosen, Ceder, Persson,
  and Jain}]{dunn2022structured}
Alexander Dunn, John Dagdelen, Nicholas Walker, Sanghoon Lee, Andrew~S Rosen,
  Gerbrand Ceder, Kristin Persson, and Anubhav Jain. 2022.
\newblock Structured information extraction from complex scientific text with
  fine-tuned large language models.
\newblock \emph{arXiv preprint arXiv:2212.05238}.

\bibitem[{Du{\v{s}}ek et~al.(2020)Du{\v{s}}ek, Novikova, and
  Rieser}]{duvsek2020evaluating}
Ond{\v{r}}ej Du{\v{s}}ek, Jekaterina Novikova, and Verena Rieser. 2020.
\newblock Evaluating the state-of-the-art of end-to-end natural language
  generation: The e2e nlg challenge.
\newblock \emph{Computer Speech \& Language}, 59:123--156.

\bibitem[{Fisch et~al.(2019)Fisch, Talmor, Jia, Seo, Choi, and
  Chen}]{fisch2019mrqa}
Adam Fisch, Alon Talmor, Robin Jia, Minjoon Seo, Eunsol Choi, and Danqi Chen.
  2019.
\newblock {MRQA} 2019 shared task: Evaluating generalization in reading
  comprehension.
\newblock In \emph{Proceedings of 2nd Machine Reading for Reading Comprehension
  (MRQA) Workshop at EMNLP}.

\bibitem[{Friedman et~al.(2021)Friedman, Dodge, and Chen}]{friedman2021single}
Dan Friedman, Ben Dodge, and Danqi Chen. 2021.
\newblock Single-dataset experts for multi-dataset qa.
\newblock In \emph{Empirical Methods in Natural Language Processing (EMNLP)}.

\bibitem[{Frosst et~al.(2019)Frosst, Papernot, and Hinton}]{frosst2019snn}
Nicholas Frosst, Nicolas Papernot, and Geoffrey~E. Hinton. 2019.
\newblock Analyzing and improving representations with the soft nearest
  neighbor loss.
\newblock In \emph{International Conference on Machine Learning}.

\bibitem[{Gravina et~al.(2018)Gravina, Rossetto, Severini, and
  Attardi}]{Gravina2018CrossAF}
Alessio Gravina, Federico Rossetto, Silvia Severini, and Giuseppe Attardi.
  2018.
\newblock Cross attention for selection-based question answering.
\newblock In \emph{NL4AI@AI*IA}.

\bibitem[{Hartigan and Wong(1979)}]{hartigan1979algorithm}
John~A Hartigan and Manchek~A Wong. 1979.
\newblock Algorithm as 136: A k-means clustering algorithm.
\newblock \emph{Journal of the royal statistical society. series c (applied
  statistics)}, 28(1):100--108.

\bibitem[{Hermann et~al.(2015)Hermann, Kočiský, Grefenstette, Espeholt, Kay,
  Suleyman, and Blunsom}]{https://doi.org/10.48550/arxiv.1506.03340}
Karl~Moritz Hermann, Tomáš Kočiský, Edward Grefenstette, Lasse Espeholt,
  Will Kay, Mustafa Suleyman, and Phil Blunsom. 2015.
\newblock \href {https://doi.org/10.48550/ARXIV.1506.03340} {Teaching machines
  to read and comprehend}.

\bibitem[{Hewlett et~al.(2016)Hewlett, Lacoste, Jones, Polosukhin, Fandrianto,
  Han, Kelcey, and Berthelot}]{hewlett2016wikireading}
Daniel Hewlett, Alexandre Lacoste, Llion Jones, Illia Polosukhin, Andrew
  Fandrianto, Jay Han, Matthew Kelcey, and David Berthelot. 2016.
\newblock \href {https://doi.org/10.18653/v1/P16-1145} {{W}iki{R}eading: A
  novel large-scale language understanding task over {W}ikipedia}.
\newblock In \emph{Proceedings of the 54th Annual Meeting of the Association
  for Computational Linguistics (Volume 1: Long Papers)}, pages 1535--1545,
  Berlin, Germany. Association for Computational Linguistics.

\bibitem[{Iyer et~al.(2023)Iyer, Vu, Moschitti, and
  Sun}]{iyer-etal-2023-question}
Roshni Iyer, Thuy Vu, Alessandro Moschitti, and Yizhou Sun. 2023.
\newblock \href {https://doi.org/10.18653/v1/2023.eacl-main.68}
  {Question-answer sentence graph for joint modeling answer selection}.
\newblock In \emph{Proceedings of the 17th Conference of the European Chapter
  of the Association for Computational Linguistics}, pages 968--979, Dubrovnik,
  Croatia. Association for Computational Linguistics.

\bibitem[{Izacard and Grave(2021)}]{izacard2021fid}
Gautier Izacard and Edouard Grave. 2021.
\newblock Leveraging passage retrieval with generative models for open domain
  question answering.
\newblock In \emph{EACL}.

\bibitem[{Joshi et~al.(2020)Joshi, Chen, Liu, Weld, Zettlemoyer, and
  Levy}]{joshi-etal-2020-spanbert}
Mandar Joshi, Danqi Chen, Yinhan Liu, Daniel~S. Weld, Luke Zettlemoyer, and
  Omer Levy. 2020.
\newblock \href {https://doi.org/10.1162/tacl_a_00300} {{S}pan{BERT}: Improving
  pre-training by representing and predicting spans}.
\newblock \emph{Transactions of the Association for Computational Linguistics},
  8:64--77.

\bibitem[{Khattab et~al.(2022)Khattab, Santhanam, Li, Hall, Liang, Potts, and
  Zaharia}]{khattab2022demonstrate}
Omar Khattab, Keshav Santhanam, Xiang~Lisa Li, David Hall, Percy Liang,
  Christopher Potts, and Matei Zaharia. 2022.
\newblock Demonstrate-search-predict: Composing retrieval and language models
  for knowledge-intensive nlp.
\newblock \emph{arXiv preprint arXiv:2212.14024}.

\bibitem[{Khosla et~al.(2020)Khosla, Teterwak, Wang, Sarna, Tian, Isola,
  Maschinot, Liu, and Krishnan}]{khosla2020supervised}
Prannay Khosla, Piotr Teterwak, Chen Wang, Aaron Sarna, Yonglong Tian, Phillip
  Isola, Aaron Maschinot, Ce~Liu, and Dilip Krishnan. 2020.
\newblock Supervised contrastive learning.
\newblock \emph{Advances in Neural Information Processing Systems},
  33:18661--18673.

\bibitem[{Kolodner(1983)}]{kolodner1983maintaining}
Janet~L. Kolodner. 1983.
\newblock \href {https://doi.org/https://doi.org/10.1016/S0364-0213(83)80001-9}
  {Maintaining organization in a dynamic long-term memory}.
\newblock \emph{Cognitive Science}, 7(4):243--280.

\bibitem[{Kočiský et~al.(2017)Kočiský, Schwarz, Blunsom, Dyer, Hermann,
  Melis, and Grefenstette}]{https://doi.org/10.48550/arxiv.1712.07040}
Tomáš Kočiský, Jonathan Schwarz, Phil Blunsom, Chris Dyer, Karl~Moritz
  Hermann, Gábor Melis, and Edward Grefenstette. 2017.
\newblock \href {https://doi.org/10.48550/ARXIV.1712.07040} {The narrativeqa
  reading comprehension challenge}.

\bibitem[{Krishnamurthy et~al.(2012)Krishnamurthy, Balakrishnan, Xu, and
  Singh}]{krishna2012efficient}
Akshay Krishnamurthy, Sivaraman Balakrishnan, Min Xu, and Aarti Singh. 2012.
\newblock Efficient active algorithms for hierarchical clustering.
\newblock In \emph{Proceedings of the 29th International Coference on
  International Conference on Machine Learning}, ICML'12, page 267–274,
  Madison, WI, USA. Omnipress.

\bibitem[{Kwiatkowski et~al.(2019)Kwiatkowski, Palomaki, Redfield, Collins,
  Parikh, Alberti, Epstein, Polosukhin, Kelcey, Devlin, Lee, Toutanova, Jones,
  Chang, Dai, Uszkoreit, Le, and Petrov}]{47761}
Tom Kwiatkowski, Jennimaria Palomaki, Olivia Redfield, Michael Collins, Ankur
  Parikh, Chris Alberti, Danielle Epstein, Illia Polosukhin, Matthew Kelcey,
  Jacob Devlin, Kenton Lee, Kristina~N. Toutanova, Llion Jones, Ming-Wei Chang,
  Andrew Dai, Jakob Uszkoreit, Quoc Le, and Slav Petrov. 2019.
\newblock Natural questions: a benchmark for question answering research.
\newblock \emph{Transactions of the Association of Computational Linguistics}.

\bibitem[{Lai et~al.(2017)Lai, Xie, Liu, Yang, and Hovy}]{lai2017large}
Guokun Lai, Qizhe Xie, Hanxiao Liu, Yiming Yang, and Eduard Hovy. 2017.
\newblock Race: Large-scale reading comprehension dataset from examinations.
\newblock \emph{arXiv preprint arXiv:1704.04683}.

\bibitem[{Lan et~al.(2019)Lan, Chen, Goodman, Gimpel, Sharma, and
  Soricut}]{DBLP:journals/corr/abs-1909-11942}
Zhenzhong Lan, Mingda Chen, Sebastian Goodman, Kevin Gimpel, Piyush Sharma, and
  Radu Soricut. 2019.
\newblock \href {http://arxiv.org/abs/1909.11942} {{ALBERT:} {A} lite {BERT}
  for self-supervised learning of language representations}.
\newblock \emph{CoRR}, abs/1909.11942.

\bibitem[{Leake(1996)}]{leake1996case}
David~B Leake. 1996.
\newblock Case-based reasoning: experiences, lessons, and future directions.

\bibitem[{Levy et~al.(2017)Levy, Seo, Choi, and
  Zettlemoyer}]{levy-etal-2017-zero}
Omer Levy, Minjoon Seo, Eunsol Choi, and Luke Zettlemoyer. 2017.
\newblock \href {https://doi.org/10.18653/v1/K17-1034} {Zero-shot relation
  extraction via reading comprehension}.
\newblock In \emph{Proceedings of the 21st Conference on Computational Natural
  Language Learning ({C}o{NLL} 2017)}, pages 333--342, Vancouver, Canada.
  Association for Computational Linguistics.

\bibitem[{Lewis et~al.(2020)Lewis, Perez, Piktus, Petroni, Karpukhin, Goyal,
  K{\"u}ttler, Lewis, Yih, Rockt{\"a}schel et~al.}]{lewis2020rag}
Patrick Lewis, Ethan Perez, Aleksandra Piktus, Fabio Petroni, Vladimir
  Karpukhin, Naman Goyal, Heinrich K{\"u}ttler, Mike Lewis, Wen-tau Yih, Tim
  Rockt{\"a}schel, et~al. 2020.
\newblock Retrieval-augmented generation for knowledge-intensive nlp tasks.
\newblock In \emph{Neurips}.

\bibitem[{Lewkowycz et~al.(2022)Lewkowycz, Andreassen, Dohan, Dyer,
  Michalewski, Ramasesh, Slone, Anil, Schlag, Gutman-Solo
  et~al.}]{lewkowycz2022solving}
Aitor Lewkowycz, Anders Andreassen, David Dohan, Ethan Dyer, Henryk
  Michalewski, Vinay Ramasesh, Ambrose Slone, Cem Anil, Imanol Schlag, Theo
  Gutman-Solo, et~al. 2022.
\newblock Solving quantitative reasoning problems with language models.
\newblock \emph{arXiv preprint arXiv:2206.14858}.

\bibitem[{Li et~al.(2015)Li, Galley, Brockett, Gao, and
  Dolan}]{li2015diversity}
Jiwei Li, Michel Galley, Chris Brockett, Jianfeng Gao, and Bill Dolan. 2015.
\newblock A diversity-promoting objective function for neural conversation
  models.
\newblock \emph{arXiv preprint arXiv:1510.03055}.

\bibitem[{Li et~al.(2023)Li, Ma, Zhuang, Gu, Su, and Chen}]{li2023few}
Tianle Li, Xueguang Ma, Alex Zhuang, Yu~Gu, Yu~Su, and Wenhu Chen. 2023.
\newblock Few-shot in-context learning for knowledge base question answering.
\newblock \emph{arXiv preprint arXiv:2305.01750}.

\bibitem[{Liu et~al.(2021)Liu, Shen, Zhang, Dolan, Carin, and
  Chen}]{liu2021makes}
Jiachang Liu, Dinghan Shen, Yizhe Zhang, Bill Dolan, Lawrence Carin, and Weizhu
  Chen. 2021.
\newblock What makes good in-context examples for gpt-$3 $?
\newblock \emph{arXiv preprint arXiv:2101.06804}.

\bibitem[{Liu et~al.(2019)Liu, Ott, Goyal, Du, Joshi, Chen, Levy, Lewis,
  Zettlemoyer, and Stoyanov}]{liu2019roberta}
Yinhan Liu, Myle Ott, Naman Goyal, Jingfei Du, Mandar Joshi, Danqi Chen, Omer
  Levy, Mike Lewis, Luke Zettlemoyer, and Veselin Stoyanov. 2019.
\newblock Roberta: A robustly optimized bert pretraining approach.
\newblock \emph{arXiv preprint arXiv:1907.11692}.

\bibitem[{Min et~al.(2022)Min, Lyu, Holtzman, Artetxe, Lewis, Hajishirzi, and
  Zettlemoyer}]{min2022rethinking}
Sewon Min, Xinxi Lyu, Ari Holtzman, Mikel Artetxe, Mike Lewis, Hannaneh
  Hajishirzi, and Luke Zettlemoyer. 2022.
\newblock Rethinking the role of demonstrations: What makes in-context learning
  work?
\newblock \emph{arXiv preprint arXiv:2202.12837}.

\bibitem[{Min et~al.(2018)Min, Zhong, Socher, and
  Xiong}]{https://doi.org/10.48550/arxiv.1805.08092}
Sewon Min, Victor Zhong, Richard Socher, and Caiming Xiong. 2018.
\newblock \href {https://doi.org/10.48550/ARXIV.1805.08092} {Efficient and
  robust question answering from minimal context over documents}.

\bibitem[{Pasupat et~al.(2021)Pasupat, Zhang, and
  Guu}]{pasupat-etal-2021-controllable}
Panupong Pasupat, Yuan Zhang, and Kelvin Guu. 2021.
\newblock \href {https://doi.org/10.18653/v1/2021.emnlp-main.607} {Controllable
  semantic parsing via retrieval augmentation}.
\newblock In \emph{Proceedings of the 2021 Conference on Empirical Methods in
  Natural Language Processing}, pages 7683--7698, Online and Punta Cana,
  Dominican Republic. Association for Computational Linguistics.

\bibitem[{Ribeiro et~al.(2016)Ribeiro, Singh, and Guestrin}]{ribeiro2016why}
Marco~Tulio Ribeiro, Sameer Singh, and Carlos Guestrin. 2016.
\newblock \href {https://doi.org/10.1145/2939672.2939778} {"why should i trust
  you?": Explaining the predictions of any classifier}.
\newblock In \emph{Proceedings of the 22nd ACM SIGKDD International Conference
  on Knowledge Discovery and Data Mining}, KDD '16, page 1135–1144, New York,
  NY, USA. Association for Computing Machinery.

\bibitem[{Richardson et~al.(2013)Richardson, Burges, and
  Renshaw}]{richardson-etal-2013-mctest}
Matthew Richardson, Christopher~J.C. Burges, and Erin Renshaw. 2013.
\newblock \href {https://aclanthology.org/D13-1020} {{MCT}est: A challenge
  dataset for the open-domain machine comprehension of text}.
\newblock In \emph{Proceedings of the 2013 Conference on Empirical Methods in
  Natural Language Processing}, pages 193--203, Seattle, Washington, USA.
  Association for Computational Linguistics.

\bibitem[{Rissland(1983)}]{rissland1983examples}
Edwina~L Rissland. 1983.
\newblock Examples in legal reasoning: Legal hypotheticals.
\newblock In \emph{IJCAI}, pages 90--93.

\bibitem[{Rubin et~al.(2021)Rubin, Herzig, and Berant}]{rubin2021learning}
Ohad Rubin, Jonathan Herzig, and Jonathan Berant. 2021.
\newblock Learning to retrieve prompts for in-context learning.
\newblock \emph{arXiv preprint arXiv:2112.08633}.

\bibitem[{Salakhutdinov and Hinton(2007)}]{salakhutdinov2007snn}
Ruslan Salakhutdinov and Geoff Hinton. 2007.
\newblock \href {https://proceedings.mlr.press/v2/salakhutdinov07a.html}
  {Learning a nonlinear embedding by preserving class neighbourhood structure}.
\newblock In \emph{Proceedings of the Eleventh International Conference on
  Artificial Intelligence and Statistics}, volume~2 of \emph{Proceedings of
  Machine Learning Research}, pages 412--419, San Juan, Puerto Rico. PMLR.

\bibitem[{Schank(1983)}]{schank1983dynamic}
Roger~C. Schank. 1983.
\newblock \emph{Dynamic Memory: A Theory of Reminding and Learning in Computers
  and People}.
\newblock Cambridge University Press, USA.

\bibitem[{Seo et~al.(2016)Seo, Kembhavi, Farhadi, and
  Hajishirzi}]{https://doi.org/10.48550/arxiv.1611.01603}
Minjoon Seo, Aniruddha Kembhavi, Ali Farhadi, and Hannaneh Hajishirzi. 2016.
\newblock \href {https://doi.org/10.48550/ARXIV.1611.01603} {Bidirectional
  attention flow for machine comprehension}.

\bibitem[{Seo et~al.(2019)Seo, Lee, Kwiatkowski, Parikh, Farhadi, and
  Hajishirzi}]{denspi}
Minjoon Seo, Jinhyuk Lee, Tom Kwiatkowski, Ankur~P Parikh, Ali Farhadi, and
  Hannaneh Hajishirzi. 2019.
\newblock Real-time open-domain question answering with dense-sparse phrase
  index.
\newblock In \emph{ACL}.

\bibitem[{Seonwoo et~al.(2020{\natexlab{a}})Seonwoo, Kim, Ha, and
  Oh}]{seonwoo2020blanc}
Yeon Seonwoo, Ji-Hoon Kim, Jung-Woo Ha, and Alice Oh. 2020{\natexlab{a}}.
\newblock Context-aware answer extraction in question answering.
\newblock In \emph{EMNLP}.

\bibitem[{Seonwoo et~al.(2020{\natexlab{b}})Seonwoo, Kim, Ha, and
  Oh}]{seonwoo-etal-2020-context}
Yeon Seonwoo, Ji-Hoon Kim, Jung-Woo Ha, and Alice Oh. 2020{\natexlab{b}}.
\newblock Context-aware answer extraction in question answering.
\newblock In \emph{EMNLP}.

\bibitem[{Shao et~al.(2020)Shao, Cui, Liu, Wang, and Hu}]{Shao_2020}
Nan Shao, Yiming Cui, Ting Liu, Shijin Wang, and Guoping Hu. 2020.
\newblock \href {https://doi.org/10.18653/v1/2020.emnlp-main.583} {Is graph
  structure necessary for multi-hop question answering?}
\newblock In \emph{Proceedings of the 2020 Conference on Empirical Methods in
  Natural Language Processing ({EMNLP})}. Association for Computational
  Linguistics.

\bibitem[{Soares et~al.(2019)Soares, FitzGerald, Ling, and
  Kwiatkowski}]{soares2019matching}
Livio~Baldini Soares, Nicholas FitzGerald, Jeffrey Ling, and Tom Kwiatkowski.
  2019.
\newblock Matching the blanks: Distributional similarity for relation learning.
\newblock In \emph{ACL}.

\bibitem[{Tevet and Berant(2020)}]{tevet2020evaluating}
Guy Tevet and Jonathan Berant. 2020.
\newblock Evaluating the evaluation of diversity in natural language
  generation.
\newblock \emph{arXiv preprint arXiv:2004.02990}.

\bibitem[{Thai et~al.(2022)Thai, Ravishankar, Abdelaziz, Chaudhary,
  Mihindukulasooriya, Naseem, Das, Kapanipathi, Fokoue, and
  McCallum}]{thai2022cbr}
Dung Thai, Srinivas Ravishankar, Ibrahim Abdelaziz, Mudit Chaudhary, Nandana
  Mihindukulasooriya, Tahira Naseem, Rajarshi Das, Pavan Kapanipathi, Achille
  Fokoue, and Andrew McCallum. 2022.
\newblock Cbr-ikb: A case-based reasoning approach for question answering over
  incomplete knowledge bases.
\newblock \emph{arXiv preprint arXiv:2204.08554}.

\bibitem[{Thayaparan et~al.(2020)Thayaparan, Valentino, and
  Freitas}]{thayaparan2020explainability_survey}
Mokanarangan Thayaparan, Marco Valentino, and André Freitas. 2020.
\newblock \href {https://doi.org/10.48550/ARXIV.2010.00389} {A survey on
  explainability in machine reading comprehension}.

\bibitem[{Trischler et~al.(2016)Trischler, Wang, Yuan, Harris, Sordoni,
  Bachman, and Suleman}]{trischler2016newsqa}
Adam Trischler, Tong Wang, Xingdi~(Eric) Yuan, Justin~D. Harris, Alessandro
  Sordoni, Philip Bachman, and Kaheer Suleman. 2016.
\newblock \href
  {https://www.microsoft.com/en-us/research/publication/newsqa-machine-comprehension-dataset/}
  {Newsqa: A machine comprehension dataset}.

\bibitem[{Tsatsaronis et~al.(2015)Tsatsaronis, Balikas, Malakasiotis, Partalas,
  Zschunke, Alvers, Weissenborn, Krithara, Petridis, Polychronopoulos,
  Almirantis, Pavlopoulos, Baskiotis, Gallinari, Arti{\`e}res, Ngomo, Heino,
  Gaussier, Barrio-Alvers, Schroeder, Androutsopoulos, and
  Paliouras}]{Tsatsaronis2015AnOO}
George Tsatsaronis, Georgios Balikas, Prodromos Malakasiotis, Ioannis Partalas,
  Matthias Zschunke, Michael~R. Alvers, Dirk Weissenborn, Anastasia Krithara,
  Sergios Petridis, Dimitris Polychronopoulos, Yannis Almirantis, John
  Pavlopoulos, Nicolas Baskiotis, Patrick Gallinari, Thierry Arti{\`e}res,
  Axel-Cyrille~Ngonga Ngomo, Norman Heino, {\'E}ric Gaussier, Liliana
  Barrio-Alvers, Michael Schroeder, Ion Androutsopoulos, and Georgios
  Paliouras. 2015.
\newblock An overview of the bioasq large-scale biomedical semantic indexing
  and question answering competition.
\newblock \emph{BMC Bioinformatics}, 16.

\bibitem[{Tu et~al.(2019)Tu, Huang, Wang, Huang, He, and
  Zhou}]{https://doi.org/10.48550/arxiv.1911.00484}
Ming Tu, Kevin Huang, Guangtao Wang, Jing Huang, Xiaodong He, and Bowen Zhou.
  2019.
\newblock \href {https://doi.org/10.48550/ARXIV.1911.00484} {Select, answer and
  explain: Interpretable multi-hop reading comprehension over multiple
  documents}.

\bibitem[{Ushio and Camacho-Collados(2021)}]{tner}
Asahi Ushio and Jose Camacho-Collados. 2021.
\newblock {T}-{NER}: An all-round python library for transformer-based named
  entity recognition.
\newblock In \emph{EACL (demo)}.

\bibitem[{Vaswani et~al.(2017)Vaswani, Shazeer, Parmar, Uszkoreit, Jones,
  Gomez, Kaiser, and Polosukhin}]{vaswani2017attention}
Ashish Vaswani, Noam Shazeer, Niki Parmar, Jakob Uszkoreit, Llion Jones,
  Aidan~N Gomez, {\L}ukasz Kaiser, and Illia Polosukhin. 2017.
\newblock Attention is all you need.
\newblock In \emph{Neurips}.

\bibitem[{Wei et~al.(2022)Wei, Wang, Schuurmans, Bosma, Chi, Le, and
  Zhou}]{wei2022chain}
Jason Wei, Xuezhi Wang, Dale Schuurmans, Maarten Bosma, Ed~Chi, Quoc Le, and
  Denny Zhou. 2022.
\newblock Chain of thought prompting elicits reasoning in large language
  models.
\newblock \emph{arXiv preprint arXiv:2201.11903}.

\bibitem[{Yagcioglu et~al.(2018)Yagcioglu, Erdem, Erdem, and
  Ikizler-Cinbis}]{https://doi.org/10.48550/arxiv.1809.00812}
Semih Yagcioglu, Aykut Erdem, Erkut Erdem, and Nazli Ikizler-Cinbis. 2018.
\newblock \href {https://doi.org/10.48550/ARXIV.1809.00812} {Recipeqa: A
  challenge dataset for multimodal comprehension of cooking recipes}.

\bibitem[{Yang et~al.(2015)Yang, Yih, and Meek}]{yang-etal-2015-wikiqa}
Yi~Yang, Wen-tau Yih, and Christopher Meek. 2015.
\newblock \href {https://doi.org/10.18653/v1/D15-1237} {{W}iki{QA}: A challenge
  dataset for open-domain question answering}.
\newblock In \emph{Proceedings of the 2015 Conference on Empirical Methods in
  Natural Language Processing}, pages 2013--2018, Lisbon, Portugal. Association
  for Computational Linguistics.

\bibitem[{Yang et~al.(2018)Yang, Qi, Zhang, Bengio, Cohen, Salakhutdinov, and
  Manning}]{yang-etal-2018-hotpotqa}
Zhilin Yang, Peng Qi, Saizheng Zhang, Yoshua Bengio, William Cohen, Ruslan
  Salakhutdinov, and Christopher~D. Manning. 2018.
\newblock \href {https://doi.org/10.18653/v1/D18-1259} {{H}otpot{QA}: A dataset
  for diverse, explainable multi-hop question answering}.
\newblock In \emph{Proceedings of the 2018 Conference on Empirical Methods in
  Natural Language Processing}, pages 2369--2380, Brussels, Belgium.
  Association for Computational Linguistics.

\bibitem[{Yang et~al.(2023)Yang, Du, Cambria, and Cardie}]{yang2023end}
Zonglin Yang, Xinya Du, Erik Cambria, and Claire Cardie. 2023.
\newblock End-to-end case-based reasoning for commonsense knowledge base
  completion.
\newblock In \emph{Proceedings of the 17th Conference of the European Chapter
  of the Association for Computational Linguistics}, pages 3491--3504.

\bibitem[{Ye et~al.(2019)Ye, Lin, Liu, Liu, and
  Sun}]{https://doi.org/10.48550/arxiv.1911.02170}
Deming Ye, Yankai Lin, Zhenghao Liu, Zhiyuan Liu, and Maosong Sun. 2019.
\newblock \href {https://doi.org/10.48550/ARXIV.1911.02170} {Multi-paragraph
  reasoning with knowledge-enhanced graph neural network}.

\bibitem[{Yu et~al.(2014)Yu, Hermann, Blunsom, and
  Pulman}]{https://doi.org/10.48550/arxiv.1412.1632}
Lei Yu, Karl~Moritz Hermann, Phil Blunsom, and Stephen Pulman. 2014.
\newblock \href {https://doi.org/10.48550/ARXIV.1412.1632} {Deep learning for
  answer sentence selection}.

\bibitem[{Zeng et~al.(2020)Zeng, Li, Li, Hu, and Hu}]{zeng2020mrc_survey}
Changchang Zeng, Shaobo Li, Qin Li, Jie Hu, and Jianjun Hu. 2020.
\newblock \href {https://doi.org/10.48550/ARXIV.2006.11880} {A survey on
  machine reading comprehension: Tasks, evaluation metrics and benchmark
  datasets}.

\bibitem[{Zhou et~al.(2022)Zhou, Sch{\"a}rli, Hou, Wei, Scales, Wang,
  Schuurmans, Bousquet, Le, and Chi}]{zhou2022least}
Denny Zhou, Nathanael Sch{\"a}rli, Le~Hou, Jason Wei, Nathan Scales, Xuezhi
  Wang, Dale Schuurmans, Olivier Bousquet, Quoc Le, and Ed~Chi. 2022.
\newblock Least-to-most prompting enables complex reasoning in large language
  models.
\newblock \emph{arXiv preprint arXiv:2205.10625}.

\end{thebibliography}
\bibliographystyle{acl_natbib}

\clearpage
\appendix
\section{Appendix}
\label{sec:appendix}

\begin{figure*}[t]
    \centering
    \includegraphics[width=\textwidth]{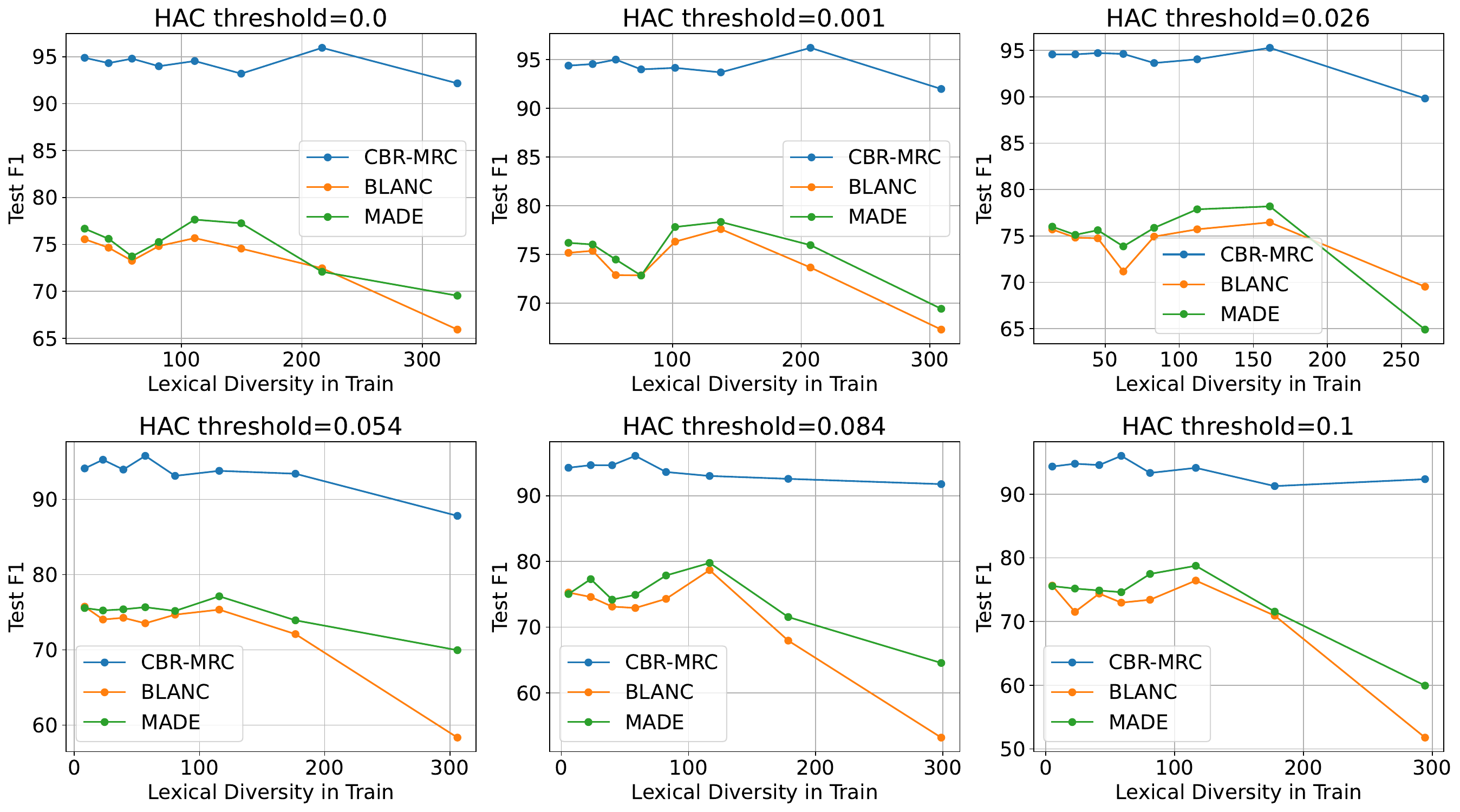}
    \caption{\textbf{F1 performance v/s lexical diversity in passage contexts on clusterings with decreasing levels of cluster-tightness.}
    We cluster questions by latent relations in the training set using HAC using 6 cut thresholds and assign each test question to one of these train clusters. We then bucket the clusters by lexical diversity scores and compute F1 performance of each bucket for each of the 6 clusterings shown.}
    \label{fig:lex-div-subplots}
\end{figure*}

\subsection{Experiments: Implementation Details}
\label{app:exp-implementation}
All \method{} models employ the BERT-base architecture and the cased vocabulary. 
We use a fixed set of hyperparameters to train all our models. 
More specifically, we use an Adam optimizer with a learning rate of $2e-5$, epsilon $1e-8$, and max gradient norm value of $5.0$.
We set the number of warm-up steps to be 8\% of the training data size. 
All models are trained for 10 epochs, which takes 2-4 days on a single RTX-8000 GPU each, depending on the size of the dataset. The standard models we report have parameter counts of 108M (BERT, MADE, SpanBERT, BLANC), 17M (ALBERT), 124M (RoBERTa), and the large variants have a parameter count of 333M.

\label{app:reuse-candidate-spans}
Our approach follows prior QA work \cite{denspi} and selects a broad set of spans that may be answers to factoid-type questions, such as entity mentions, date-time strings, numbers, and quoted strings. We also include all n-grams up to three words. For date-time string extraction, we use the $datefinder$ Python package, and for entity mentions, we use the T-NER library \cite{tner}.

\subsection{Experiments: Dataset Statistics}
\label{app:exp-datasets}
For our experiments, we use the curated versions of NaturalQuestions, NewsQA, BioASQ, and RelEx datasets from the MRQA shared task \cite{fisch2019mrqa}. Table \ref{tab:dataset_stats} presents the number of examples present in each of the datasets used in the paper.
\begin{table}[h]
\centering
    \begin{tabular}{lcc}
    \toprule
     \bf Dataset & \bf Train & \bf Dev \\
     \midrule
    NaturalQ & 104,071 & 12,836  \\
    NewsQA & 74,160 & 4,212  \\
    BioASQ & - & 1,504  \\
    RelEx & - & 2,948  \\
    \bottomrule
    \end{tabular}
\caption{Dataset statistics}
\label{tab:dataset_stats}
\end{table}


\subsection{Lexical Diversity Analysis: Distribution in NaturalQuestions}
\label{app:lex-NQ}
We evaluate the distribution of lexical diversity in the NaturalQuestions dataset. Out of the set of test questions under consideration in our case study on lexical diversity, we find that $7.46\%$, or $504$, unique (question, answer) pairs contain more than one context, which is used to extract the answer, highlighting the relevance of lexical diversity as a consideration for models to address. A question in this set may contain up to 4 different contexts that can answer the same exact question. In Table~\ref{tab:lex-div-qual}, we show examples from the test set of questions that can be answered using multiple, diverse contexts.


\subsection{Lexical Diversity Analysis: Clustering questions by latent relations}
\label{app:lex-clustering}
We cluster questions in the training set based on the latent relations they contain using hierarchical agglomerative clustering (HAC) \citep{krishna2012efficient} over masked similarity scores\footnote{We restrict the number of nearest-neighbors to 20 for each question.}. Importantly, since we do not have ground-truth latent relations for questions, we run our analysis on \textit{multiple} flat clusterings, where each cluster in a clustering should represent the same semantic relations. To obtain flat clusters in a fair manner, we cut the HAC tree using $C$ thresholds, computed using $k$-means \citep{hartigan1979algorithm} over similarity scores across all questions in the training set. This yields $C$ clusterings with increasing levels of cluster tightness, i.e., how similar each question is in a cluster. We set $C$ to 6 and lower-bound the similarity scores to $0.9$ to reduce noise in the clusters. For each clustering, we then compute the per-cluster lexical diversity as the count of the number of unique tokens in the set of passage contexts across questions in the cluster, normalized by the cluster size. Finally, we cluster the test set by assigning each question to the cluster label of its top-1 similar train question for each of the $C$ clusterings, resulting in 6 test clusterings. In our analysis, we retain only unique $(q, \answerset, p)$ triples resulting in 7,213 instances from an original set of 12,836. We additionally drop 98 questions that cannot be answered by any of the three models under consideration, resulting in a final set of 7,115 test questions.

\subsection{Lexical Diversity Analysis: F1 Performance}
\label{app:lex-plots}
Figure~\ref{fig:lex-div-subplots} shows the absolute F1 performance of \method{} and the two baselines (BLANC and MADE) on each clustering separately. Note that Figure~\ref{fig:lexical-diversity} is a reduction from this, in that it takes the average difference between the F1 score of \method{} and the baselines across the 6 clusters.


\subsection{Does \method{} have the correct inductive bias?}
Text representations obtained from pre-trained language models have shown the ability to capture the latent semantics of text spans. 
In this experiment, we evaluate \method{} when the parameters are set to the pre-trained BERT-base model \emph{without} fine-tuning. As shown in Table~\ref{tab:pretrained_results}, we find that our model with no fine-tuning performs surprisingly well, indicating that \method{} does have the correct inductive bias. However, fine-tuning with our proposed contrastive loss leads to a significant increase in performance.

\begin{table}[h]
\centering
\resizebox{\columnwidth}{!}{
    \begin{tabular}{lccH}
    \toprule
    \textbf{Model} & \textbf{NewsQA} & \textbf{NaturalQ} & $\mathbf{\Delta}$ \\
    \midrule
    BERT & 50.11 & 64.48 & +0.00 \\
    SpanBERT  & 52.85 & 66.60 & +2.12 \\
    \midrule
    \bf \method{}$_{\textsc{Pre-Trained}}$ & $18.63$ & $46.07$ & $-18.41$ \\
    \bf \method{}$_{\textsc{Fine-Tuned}}$ & $64.95$ & $\mathbf{82.12}$ & $+17.64$ \\
    \bottomrule
    \end{tabular}
}
\caption{\textbf{Pre-trained \method{} performance.} We report the performance (EM) of \method{} using both a pre-trained passage encoder as well as an encoder fine-tuned with the contrastive loss as described in \S\ref{sec:method}. We additionally include scores from reasonable, \textit{fine-tuned} baselines for comparison. 
}
\label{tab:pretrained_results}
\end{table}

\begin{table*}[htb]
    \small
    \centering
    \renewcommand{\arraystretch}{2} 
    \resizebox{0.9\textwidth}{!}{
    \begin{tabular}{p{0.2\linewidth} p{0.08\linewidth} p{0.72\linewidth}}
    \toprule
    
    \textbf{Question} & \textbf{Answer} & \textbf{Context} \\
    \midrule[0.1pt]
     when does season 5 of ruby come out & October 14, 2017 & [...] Four seasons , referred to as `` Volumes '' , have been released , with a fifth currently ongoing since its premiere on October 14 , 2017 . As of [...] \\
    &  & [...] The fifth Volume premiered on October 14 , 2017 , a date which was first announced at the RTX Austin 2017 event . The episodes [...] \\
    &  & [...] <Table> <Tr> <Th colspan="2"> Season </Th> <Th colspan="2"> Episodes </Th> <Th colspan="2"> Originally aired </Th> [...] <Td colspan="1"> 5 </Td> <Td colspan="2"> 14 </Td> <Td colspan="1"> October 14 , 2017 ( 2017 - 10 - 14 ) </Td> <Td> [...]\\
    
    \midrule[0.1pt]
     who wrote most of the declaration of independence & Thomas Jefferson & [...] Committee of Five '' to draft a declaration , consisting of John Adams of Massachusetts , Benjamin Franklin of Pennsylvania , Thomas Jefferson of Virginia , Robert R. Livingston of New York , and Roger Sherman of Connecticut . The committee [...] \\
    &  & [...] John Adams , a leader in pushing for independence , had persuaded the committee to select Thomas Jefferson to compose the original draft of the document , which Congress edited to produce the final version [...] \\
    &  & [...] The source copy used for this printing has been lost and may have been a copy in Thomas Jefferson 's hand . Jefferson 's original draft is preserved at the Library of Congress , complete with changes made by John Adams [...] \\

    \midrule[0.1pt]
     who plays the mom on the tv show mom & Allison Janney & [...] working as a waitress and attending Alcoholics Anonymous meetings . Her mother Bonnie Plunkett ( Allison Janney ) is also a recovering drug and alcohol addict . Christy 's daughter [...] \\
    &  & [...] It stars Anna Faris and Allison Janney in lead roles as dysfunctional daughter / mother duo Christy and Bonnie Plunkett [...] \\
    &  & [...] her relationship with her mother Bonnie [...] <Li> Allison Janney as Bonnie Plunkett : Christy 's mother , a joyful if cynical recovering addict who is now grateful with life . She tries [...] \\

    \midrule[0.1pt]
     when was the us department of homeland security created & November 25 , 2002 & <Tr> <Th> Formed </Th> <Td> November 25 , 2002 ; 15 years ago ( 2002 - 11 - 25 ) </Td> </Tr> \\
    &  & [...] <Th colspan="2"> Agency overview </Th> </Tr> <Tr> <Th> Formed </Th> <Td> November 25 , 2002 ; 15 years ago ( 2002 - 11 - 25 ) </Td> [...] \\
    &  & [...] The Department of Homeland Security was established on November 25 , 2002 , by the Homeland Security Act of 2002 . It was intended to consolidate U.S. [...] \\

    \midrule[0.1pt]
     who has won the eurovision song contest the most times & Ireland & [...] Ireland has finished first seven times , more than any other country [...] \\
    &  & [...] <Table> <Tr> <Th> Wins </Th> <Th> Country </Th> <Th> Years </Th> </Tr> <Tr> <Td> 7 </Td> <Td> Ireland </Td> <Td> 1970 , 1980 , 1987 , 1992 , 1993 , 1994 , 1996 </Td> </Tr> <Tr> <Td> 6 </Td> <Td> Sweden </Td> <Td> 1974 , 1984 , 1991 , 1999 , 2012 , 2015 </Td> </Tr> <Tr> <Td> [...] \\
    &  & [...] The country with the highest number of wins is Ireland , with seven . The only person to have won more than once [...] \\
    
    \bottomrule
    \end{tabular}
    }
    \caption{\textbf{Lexical Diversity in NaturalQuestions.} We show here unique question-answer pairs that can be answered using diverse contexts.}
    \label{tab:lex-div-qual}
\end{table*}

\end{document}